\pdfoutput=1

\documentclass[journal]{IEEEtran}

\newif\ifanonymous
\anonymousfalse
\newif\ifarxiv     %
\arxivtrue

\IEEEoverridecommandlockouts

\usepackage{cite}
\usepackage{amsmath, amssymb, amsfonts}
\usepackage{graphicx}
\graphicspath{{figures/}}
\usepackage{textcomp}
\usepackage{xcolor}
\usepackage{booktabs}
\usepackage{array}
\usepackage{multirow}
\usepackage{siunitx}
\usepackage[caption=false, font=footnotesize]{subfig}   %
\usepackage{balance}
\usepackage{xspace}
\usepackage{url}
\usepackage{capt-of}                                     %
\usepackage{listings}                                     %
\lstdefinestyle{prompt}{basicstyle=\fontencoding{T1}\fontfamily{lmtt}\selectfont\scriptsize, breaklines=true, breakatwhitespace=false,
  columns=fullflexible, keepspaces=true, showstringspaces=false, upquote=true, frame=single, framerule=0.4pt,
  rulecolor=\color{black!35}, backgroundcolor=\color{black!3}, xleftmargin=3pt, xrightmargin=3pt, framesep=3pt,
  aboveskip=3pt, belowskip=8pt, inputencoding=utf8, extendedchars=true,
  literate={—}{{{\rmfamily\textemdash}}}1 {·}{{\textperiodcentered}}1 {°}{{\textdegree}}1 {~}{{\textasciitilde}}1 {-}{{-}}1}
\newcommand{\promptfile}[2]{\par\noindent\textbf{#1}\par\nopagebreak\lstinputlisting[style=prompt]{#2}}
\makeatletter

\long\def\@makecaption#1#2{%
\ifx\@captype\@IEEEtablestring%
\setbox\@tempboxa\hbox{\normalfont\footnotesize {#1.}\nobreakspace\nobreakspace #2}%
\ifdim \wd\@tempboxa >\hsize%
\setbox\@tempboxa\hbox{\normalfont\footnotesize {#1.}\nobreakspace\nobreakspace}%
\parbox[t]{\hsize}{\normalfont\footnotesize\noindent\unhbox\@tempboxa#2}%
\else\hbox to\hsize{\normalfont\footnotesize\box\@tempboxa\hfil}\fi
\par\vskip 3pt     %
\else
\@IEEEfigurecaptionsepspace
\setbox\@tempboxa\hbox{\normalfont\footnotesize {#1.}\nobreakspace\nobreakspace #2}%
\ifdim \wd\@tempboxa >\hsize%
\setbox\@tempboxa\hbox{\normalfont\footnotesize {#1.}\nobreakspace\nobreakspace}%
\parbox[t]{\hsize}{\normalfont\footnotesize\noindent\unhbox\@tempboxa#2}%
\else\hbox to\hsize{\normalfont\footnotesize\box\@tempboxa\hfil}\fi\fi}
\makeatother

\usepackage[colorlinks=true, allcolors=blue!50!black, breaklinks=true, bookmarks=false]{hyperref}
\usepackage[capitalize, noabbrev]{cleveref}
\crefname{appendix}{App.~Sec.}{App.~Secs.}\Crefname{appendix}{App.~Sec.}{App.~Secs.}
\crefname{subappendix}{App.~Sec.}{App.~Secs.}\Crefname{subappendix}{App.~Sec.}{App.~Secs.}

\newcommand{\method}{WayTrace\xspace}

\newcommand{\anon}[2]{\ifanonymous#1\else#2\fi}
\newcommand{\cityname}{\anon{City~X}{Zurich}\xspace}
\newcommand{\imagery}{\anon{}{ Imagery \textcopyright{}~swisstopo.}}
\newcommand{\anonfig}[1]{#1\ifanonymous _anon\fi}

\newcommand{\vect}[1]{\mathbf{#1}}
\newcommand{\norm}[1]{\left\lVert #1 \right\rVert}

\makeatletter\let\tabinput\@@input\makeatother
\newcommand{\up}{$\uparrow$}
\newcommand{\down}{$\downarrow$}

\begin{document}
\bstctlcite{IEEEexample:BSTcontrol}   %

\title{\method: A Map-to-Language-to-Trace Hierarchy\\for Embodied Urban Navigation}

  \author{Manthan Patel$^{1*}$, Jonas Frey$^{2}$, Pascal Roth$^{1, 3}$, Marco Hutter$^{1}$, and Cesar Cadena$^{1}$
\thanks{This work has been submitted to the IEEE for possible publication. Copyright may be transferred without
notice, after which this version may no longer be accessible.}   %
\thanks{$^1$ Robotic Systems Lab (RSL), ETH Zurich, Switzerland}
\thanks{$^2$ UC Berkeley, Berkeley, USA}
\thanks{$^3$ NVIDIA, Zurich, Switzerland}
\thanks{$^*$ Corresponding: \texttt{patelm@ethz.ch}}}

\IEEEaftertitletext{\vspace{-0.5\baselineskip}
\begin{center}
  \includegraphics[width=\textwidth]{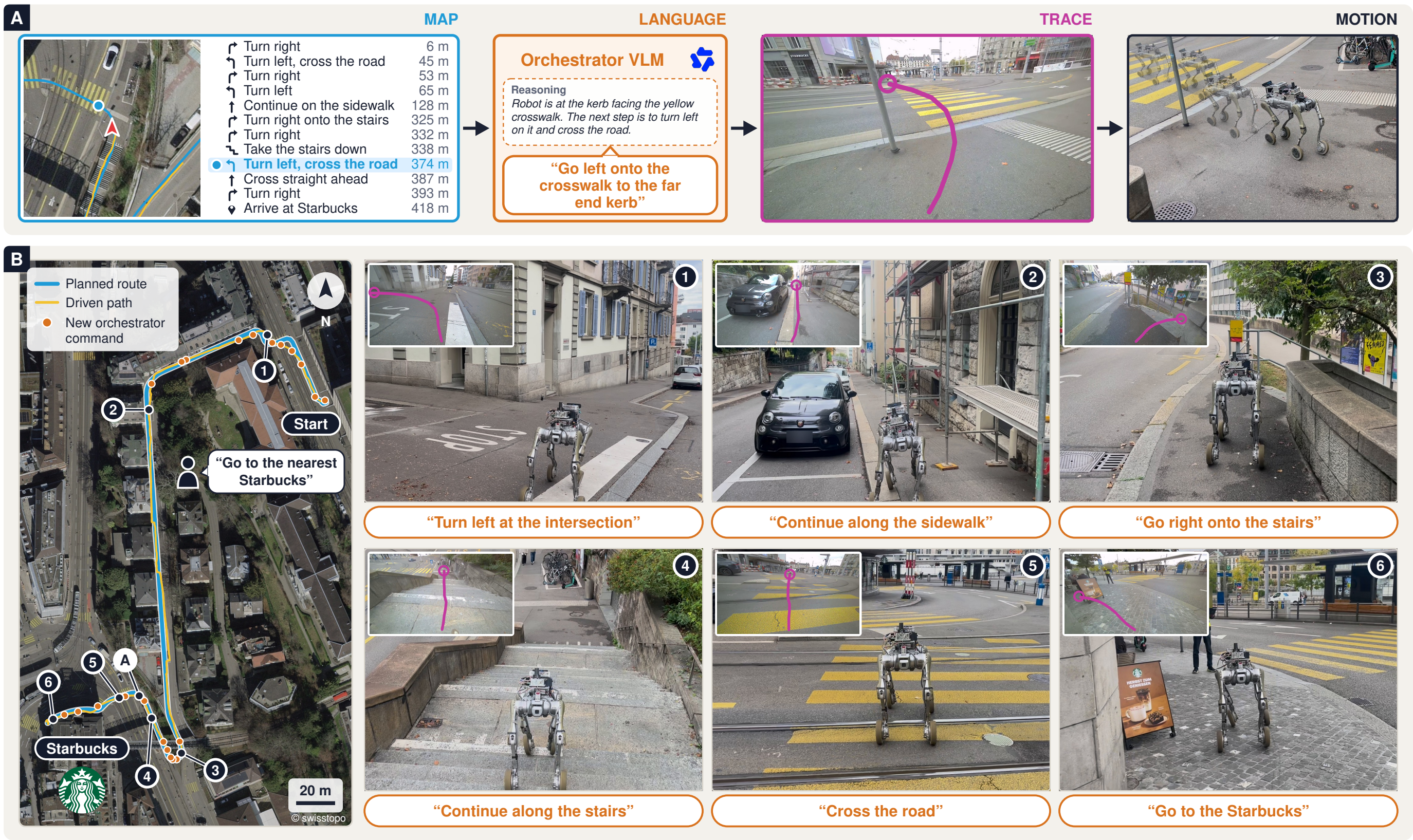}
  \captionof{figure}{Real-world deployment of \method through a city centre, given the request ``Go to the nearest
  Starbucks'' (426\,m away). (A)~One step of the hierarchy at the entry to a crossing. A web map (e.g., Google Maps) provides
  turn-by-turn language directions once at the start of the mission (\textsc{Map}, shortened here). The orchestrator VLM
  (\textsc{Language}, Qwen3.8-Flash-Next on a remote server) receives these directions, a
  robot-centric top-down satellite crop, the current camera image and a history of past images, and it issues the next
  atomic language command, shown with its condensed reasoning. The trace predictor VLM (\textsc{Trace}, onboard)
  predicts an image-space trace conditioned on that command. The trace is lifted to 3D using depth and tracked by an
  RL-based navigation policy (\textsc{Motion}). (B)~The planned route, the driven path and each new orchestrator command
  (orange dots) on the satellite map, together with six moments along the mission. Each moment shows a third-person view
  of the robot, the robot camera with the predicted trace (inset) and the current orchestrator command.}
  \label{fig:teaser}
\end{center}
\vspace{0.5\baselineskip}
}

\maketitle

\begin{abstract}
Web services provide walking directions for nearly every city, but following them reliably over long routes remains
challenging for autonomous robots. Point-goal navigation policies rely heavily on accurate GNSS, while most vision-language
navigation models struggle to execute even short instructions outdoors. We present \method, a hierarchical system that connects
web-map directions to robot motion through language. An off-the-shelf orchestrator vision-language model reads the
turn-by-turn directions, a satellite map and recent camera images, and issues one atomic language command at a time, such as
``cross the road to the left''. An onboard trace predictor, trained on over one million automatically labelled samples from street-level images, draws the path for this command in the image, and a reinforcement learning policy tracks it. In long-range
simulated missions, \method remains robust to GNSS noise and rarely leaves crosswalks or sidewalks, while point-goal
baselines degrade sharply with about ten times more violations. Deployed on a wheeled-legged robot, \method completes
real-world missions of up to 1.37\,km at about 1.1\,m/s through a busy city centre, rural trails and a university
campus, handling crosswalks, staircases and construction sites.
 Project page: \url{https://way-trace.github.io}   %

\end{abstract}

\begin{IEEEkeywords}
Navigation, Vision-Language-Action Models, Foundation Models, Embodied AI
\end{IEEEkeywords}

\section{Introduction}
\label{sec:intro}

\IEEEPARstart{H}{umans} navigate unfamiliar cities with little effort. Given a destination, a person consults a web map
and follows its turn-by-turn directions, such as continuing along a sidewalk, turning onto a crosswalk or taking the stairs. Neither a metric goal nor precise position is required. The map indicates the next manoeuvre and its
approximate location, while vision resolves where the crosswalk begins and when a turn has been completed.
Since web maps provide such directions for nearly every place where people walk, a robot that follows them in the same manner
could be deployed in any city given only a destination and without a prior mapping run, enabling applications such as
last-mile delivery, urban inspection and guidance of visually impaired people.

Existing learned navigation methods fall short of this goal over long routes.
Point-goal methods~\cite{roth2024viplanner,liu2025citywalker,castro2025vamos,hirose2025omnivla,yang2025sru} receive the route as a sequence of GNSS waypoints, and their performance is therefore bounded by the accuracy of the position estimate.
In cities, GNSS errors of several metres are common~\cite{merry2019smartphone}, enough to place a waypoint on the road or on the opposite side of the street, and over routes of several hundred metres such errors inevitably lead to failures.
Waypoints also carry no notion of the social norms, such as staying on the sidewalk and crossing at crosswalks, so a policy that reaches every waypoint may still violate them and pose a safety risk in urban traffic.

Vision-language navigation (VLN) replaces coordinates with natural-language instructions,
but it has been studied predominantly in simulated indoor environments~\cite{anderson2018r2r,ku2020rxr} and on street-view panoramas~\cite{chen2019touchdown}.
Recent models bring vision-language navigation to real
robots~\cite{huang2026ticvla,zhang2026foresight,cheng2025navila,zhang2025ventura}, but they are typically demonstrated on single instructions over short distances.
We find that these models are brittle even at the level of a single atomic
instruction (\cref{sec:q4}), which makes it infeasible to follow a kilometre-long route by chaining such instructions.

In this work, we simplify the problem. Web-map directions already decompose a route into a sequence of atomic instructions, each describing a single manoeuvre such as ``cross the road'' that can be grounded in the camera image.
Long-range navigation thus separates into deciding which atomic instruction applies at the current moment and executing that instruction reliably.
The first problem requires the map and contextual reasoning but only a coarse position, since GNSS is needed only to identify the current step of the route rather than the exact position of the robot.
The second problem requires only the camera and a short observation history.
Language is a suitable interface between the two
levels, since it is what web maps produce and what vision-language models (VLMs) reason in.
Based on this observation, we propose \method, a map-to-language-to-trace hierarchy (\cref{fig:teaser}).
An off-the-shelf orchestrator VLM reads the web-map directions, a satellite map of the surroundings and the recent camera images, and issues the next atomic
command. An onboard trace predictor draws the path for this command in image space, and a reinforcement learning (RL) policy tracks the path lifted to 3D.

We evaluate \method on ten long-range missions in a simulated \cityname old town and in real-world deployments. With
perfect GNSS, \method completes 83\,\% of the missions. It achieves the highest success rate at every level of GNSS
error, and its success rate decreases gradually to 63\,\% with a drifting GNSS error of 7\,m RMS, whereas that of point-goal
baselines drops below 25\,\% and they leave the crosswalk or the sidewalk about ten times as often. Vision-language baselines complete
at most 7\,\% of the missions, even with language commands from a tuned oracle. In the real world, \method completed
three missions of up to 426\,m through a busy city centre fully autonomously, and a 1.37\,km mission along rural trails
and through a university campus with two short operator interventions. The main contributions of this work are as
follows.
\begin{itemize}
  \item \method, a hierarchical navigation framework that translates web-map directions into atomic language commands
    and executes them as image-space traces.
  \item A language-conditioned trace predictor together with an automatic labelling pipeline that produces most of its
    training data from unlabelled street-level image sequences.
  \item An extensive evaluation against point-goal and vision-language navigation baselines in simulation, an analysis of each
    component and real-world deployments on a wheeled-legged robot.
\end{itemize}
We will release the simulation environment, the long-range navigation benchmark, the trace prediction benchmarks, the
dataset, the models and the code.

\section{Related Work}
\label{sec:related}

\subsection{Long-Range and Map-Guided Outdoor Navigation}

Navigation foundation models learn goal-conditioned policies from data pooled across robots and
environments~\cite{shah2023gnm,shah2023vint,sridhar2024nomad}, and recent work scales point-goal navigation with
crowd-sourced robot data~\cite{suomela2026datascaling,hirose2026mbra} or with walking videos from the
web~\cite{liu2025citywalker,xia2026longtail}. Vision-language-action models for navigation additionally accept GNSS
poses, images or language as goals~\cite{hirose2025omnivla}. VAMOS~\cite{castro2025vamos} plans candidate paths in the
image with a VLM and re-ranks them with an embodiment-specific affordance model. Over long distances, these policies are
guided by GNSS goals or waypoints, so the route they follow is only as accurate as the position estimate.

Several systems follow routes from map services. ViKiNG~\cite{shah2022viking} reached kilometre-scale goals by using
overhead images as heuristics for a learned local policy, and Sorokin et al.~\cite{sorokin2022sidewalks} followed public
map routes with a legged robot while staying on sidewalks. UrbanVLA~\cite{li2026urbanvla} aligns noisy route waypoints
from a map service with the camera observations in an end-to-end policy. Walk With Me~\cite{zhang2026walkwithme} lets a
VLM ground instructions in map data and GNSS as waypoints for a low-level policy and defers difficult situations, such
as crossings, to the VLM. DA-Nav~\cite{yuan2026danav} reduces web-map directions to four discrete commands that a VLM
grounds as cells in the image, and NaViRrator~\cite{lee2026navirrator} verbalises a map route once into an instruction
for a VLN policy. In \method, the route is converted into atomic language commands in closed loop. Each command is
grounded in the current observation rather than in metric coordinates, and hence GNSS errors do not displace the path
that the robot follows.

\subsection{Vision-Language Navigation}

Vision-language navigation was established on simulated indoor
benchmarks~\cite{anderson2018r2r,ku2020rxr,krantz2020vlnce} and on street-view graphs of real
cities~\cite{chen2019touchdown,schumann2021map2seq,vasudevan2021talk2nav,schumann2024velma}, where an agent follows a
paragraph of instructions written for the whole route. Video-based VLMs transfer such models to real
robots~\cite{zhang2024navid,zhang2025uninavid,wei2026streamvln}, and recent navigation foundation models scale this
approach to millions of samples across tasks and embodiments, unifying instruction following, object search and point
goals in a single
model~\cite{zhang2025navfom,chu2026abotn0,gong2026abotn1,zhang2026qwenrobotnav,wang2026lightnav0,bounhar2026robostral}.
NaVILA~\cite{cheng2025navila} lets a VLM issue metric language actions that an RL locomotion policy executes on a legged
robot, and UrbanNav~\cite{mei2026urbannav} learns language-guided urban navigation from city-walking videos.
TIC-VLA~\cite{huang2026ticvla} couples language reasoning with control and accounts for the delay of its reasoning in
dynamic environments. Foresight~\cite{zhang2026foresight} lets a VLM propose motion plans in the image and iteratively
critique and refine them. VENTURA~\cite{zhang2025ventura} adapts an image diffusion model to generate path masks that a
lightweight policy executes. These models execute one instruction or goal at a time, which a user or a higher-level
agent has to provide~\cite{zhang2026qwenrobotnav}. \method keeps each instruction atomic and lets an orchestrator decide
which instruction applies along the route.

\subsection{Hierarchical Systems with Language as the Interface}

Decomposing long-horizon tasks into language subtasks for a low-level policy is well established in manipulation.
In SayCan~\cite{ahn2022saycan}, a language model proposes skills that are scored by affordance functions, and RT-H~\cite{belkhale2024rth} predicts language motions as an intermediate step between the task and the actions.
Hi Robot~\cite{shi2025hirobot} uses a high-level VLM to issue atomic commands to a vision-language-action model, while $\pi_{0.5}$~\cite{black2025pi05} uses a single model that first predicts the next subtask in language and then generates the actions for it.
In MEM~\cite{torne2026mem}, a high-level policy predicts the next subtask together with a language memory of task progress, and a low-level policy executes it with a short-term visual memory.
In navigation, LM-Nav~\cite{shah2023lmnav} parses instructions into landmarks on a previously built topological graph, and Mobility VLA~\cite{xu2024mobilityvla} selects a goal frame from a recorded tour.
NavMCP~\cite{lei2026navmcp} lets a VLM agent invoke a navigation foundation model for long-horizon exploration.
VbotNav-0~\cite{tan2026vbotnav} and AsyncVLA~\cite{hirose2026asyncvla} split navigation between a remote and an onboard policy, which exchange waypoints and
latent action features, respectively, rather than language.
\method uses language between an off-the-shelf orchestrator and an onboard trace predictor, which requires no prior visit of the environment and keeps every decision interpretable.

\subsection{Image-Space Trace Prediction and Automatic Labelling}

Predicting a sequence of 2D waypoints in image space, referred to as a trace, to accomplish a navigation task has recently gained popularity with VLMs.
NaviTrace~\cite{navitrace} benchmarks VLMs on predicting such traces in the image for a given instruction and embodiment.
Recent navigation models predict either a trace~\cite{fu2026trajvln,wang2026crosstracer,castro2025vamos,zhang2026foresight,zhang2025ventura} or only the next
waypoint as a point in the image~\cite{wang2026lightnav0,bounhar2026robostral}. Following this line of work, \method uses image-space traces as the interface between the language command and the low-level controller, since they can be learned at scale without robot data.
Training a trace predictor, however, requires traces that match a specific instruction.
Prior work derives navigation labels from web videos~\cite{liu2025citywalker,mei2026urbannav,hirose2024lelan} or from monocular reconstruction of egocentric video~\cite{seneviratne2026vega}.
We, instead, build an automatic labelling pipeline that derives such command-specific traces from unlabelled street-level image sequences~\cite{mapillary} with 3D reconstruction~\cite{wang2026vggtomega} and VLM labelling.

\section{Method}
\label{sec:method}

\begin{figure*}[t]
\centering
\includegraphics[width=\textwidth]{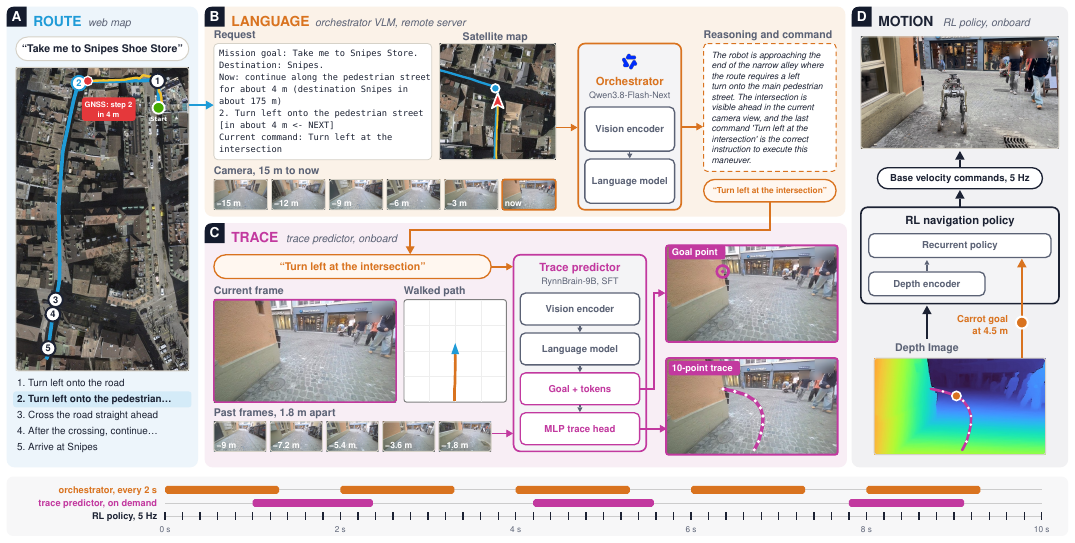}
\par\vspace{-7pt}%
\caption{Overview of \method with the logged inputs and outputs of one decision in the real-world Snipes mission
(\cref{fig:realworld}). (A)~Once per mission, the web map provides the route and turn-by-turn directions, and GNSS
only identifies the next step and its distance. (B)~Every 2\,s, the off-the-shelf orchestrator VLM on a remote server
reads the directions, a heading-up satellite map and the camera images of the last 15\,m, and it answers with its
reasoning and the next atomic command. (C)~Onboard, the trace predictor receives this command, five past frames 1.8\,m
apart, the current frame and the walked path from odometry, decodes a goal point and regresses a ten-point trace in the
current image. (D)~The trace is lifted to 3D with stereo depth, and an RL navigation policy follows a carrot 4.5\,m
ahead along it. The loops run asynchronously, and the robot keeps moving on the previous path while a new command or
trace is computed, as shown in the timeline at the bottom.}
\label{fig:overview}
\end{figure*}

\subsection{Problem Formulation and Overview}
\label{sec:problem}

A robot has to reach a destination $d$ in a city, given as text, while it moves on sidewalks and crosses roads at
crosswalks. At time $t$, it observes an RGB image $I_t$ and a depth image $D_t$ from a front camera, its velocity
$\vect{v}_t$ and pose from odometry and a GNSS position that may be several metres wrong, and it outputs base velocity
commands $\vect{a}_t$. A web map provides a walking route with turn-by-turn directions, and no metric map of the
environment is built beforehand.

Neither the route nor the GNSS position is accurate enough to be tracked geometrically. \method therefore expresses the
route as a sequence of atomic commands $c \in \mathcal{C}$. An atomic command describes a single manoeuvre whose extent
is visible from the current viewpoint, such as ``Turn left at the intersection'' or ``Walk to the bench on the right'',
without referring to positions or distances. Navigation then separates into deciding which command applies at the
current moment and executing this command. \method solves the first problem with an orchestrator VLM $\mathcal{O}$ and
the second with a trace predictor $f_\theta$ and an RL navigation policy $\pi$ (\cref{fig:overview}),
\begin{equation}
  \begin{gathered}
  c_t = \mathcal{O}\big(\vect{x}^{o}_{t-\Delta_o}\big), \qquad
  \boldsymbol{\tau}_t = f_\theta\big(\vect{x}_{t-\Delta_\tau}, c_t\big), \\
  \vect{a}_t = \pi\big(D_t, \vect{q}_t, \vect{v}_t\big),
  \end{gathered}
  \label{eq:hierarchy}
\end{equation}
where $\vect{x}^{o}$ and $\vect{x}$ are the inputs of the orchestrator and the trace predictor, $\boldsymbol{\tau}_t$ is
a trace of points in the current image, $\vect{q}_t$ is a goal point on the trace lifted to 3D, and $\Delta_o \approx
1.3$\,s and $\Delta_\tau \approx 1.5$\,s are the latencies of the two models. The levels run asynchronously, and the
policy follows the latest trace at 5\,Hz, so that the robot keeps moving while new commands and traces are computed.

\subsection{Route and Orchestrator VLM}
\label{sec:route}
\label{sec:orchestrator}

The route comes from the Google Routes API and is replanned on the pedestrian network of OpenStreetMap, so that it uses
sidewalks and mapped crossings. It is divided into directions in the style of the web map, such as ``Turn left onto the
crosswalk and cross the road'', each attached to a position along the route. The GNSS position, projected onto the
route, only selects the next direction and its approximate distance (\cref{app:implementation}).

Every 2\,s, an off-the-shelf VLM decides whether the current command still applies or the next manoeuvre is due. Its
input $\vect{x}^{o}_t$ contains the destination, the nearby directions with their distances, a heading-up satellite
image of 50\,m width with the route and the driven path, the camera images of the last 15\,m and the current command. It
answers with a short reasoning followed by the command. The prompt (\cref{app:prompts}) instructs it to determine
\emph{which} manoeuvre comes next and on which side from the directions and the map, and \emph{when} it is due from the
camera images, since distances derived from GNSS may be wrong by 10 to 20\,m. A few rules on the robot reject commands
that are issued far too early (\cref{app:implementation}).

\subsection{Language-Conditioned Trace Predictor}
\label{sec:trace}

The trace predictor executes the current command as a trace in the camera image, which can be learned without robot data
and lifted to 3D with depth. Its input $\vect{x}_t$ consists of five past images taken 1.8\,m of travel apart, the
current image $I_t$ and a top-down map $B_t$ of the path walked over the last 16\,m. The history lets the model
recognise which part of the command has already been executed, for instance that a commanded turn is complete. The
output is a goal point $\vect{g}$ and a trace $\boldsymbol{\tau} = (\vect{u}_1, \ldots, \vect{u}_N)$ of $N = 10$ points
in normalised image coordinates that ends at the goal, where the goal is the point at which the commanded path leaves
the view or becomes occluded.

The model is initialised from RynnBrain-9B~\cite{dang2026rynnbrain}, writes the coordinates as text and predicts the
goal before the trace,
\begin{equation}
  p_\theta(\boldsymbol{\tau} \mid \vect{x}, c) = p_\theta(\vect{g} \mid \vect{x}, c)\,
  p_\theta(\boldsymbol{\tau} \mid \vect{g}, \vect{x}, c).
  \label{eq:twostage}
\end{equation}
Predicting the goal first commits the model to a single branch at a junction, whereas a model that predicts the trace
directly tends to hedge between branches. Both stages are trained by supervised fine-tuning with the autoregressive
cross-entropy loss, the second stage with the reference goal in the prompt.

For deployment, a single call replaces the two stages. The goal of about 15 tokens is decoded autoregressively, and an
MLP head $h_\phi$ regresses the trace $\hat{\boldsymbol{\tau}} = (\hat{\vect{u}}_1, \ldots, \hat{\vect{u}}_N) =
\sigma(h_\phi(\vect{e}))$ from the hidden state $\vect{e}$ at the end of the goal, which avoids decoding about 140 trace tokens.
The head is trained with
\begin{equation}
  \mathcal{L} = \frac{1}{N}\sum_{i=1}^{N} \norm{\hat{\vect{u}}_i - \vect{u}_i}_1
  + \lambda_{g} \norm{\hat{\vect{u}}_N - \vect{g}}_1 + \lambda_{\mathrm{CE}}\, \mathcal{L}_{\mathrm{CE}},
  \label{eq:head}
\end{equation}
where $\vect{u}_i$ are the points of the reference trace, the second term adds weight to the end point, which is the
goal, $\mathcal{L}_{\mathrm{CE}}$ is the cross-entropy on the goal tokens, and $\lambda_g$ and $\lambda_{\mathrm{CE}}$
are weights (\cref{app:implementation}). The single-call model performs on par with
the two-stage model and runs onboard the robot (\cref{sec:impl}).

\subsection{Training Data with Intent-Conditioned Traces}
\label{sec:data}

\begin{figure*}[t]
\centering
\includegraphics[width=\textwidth]{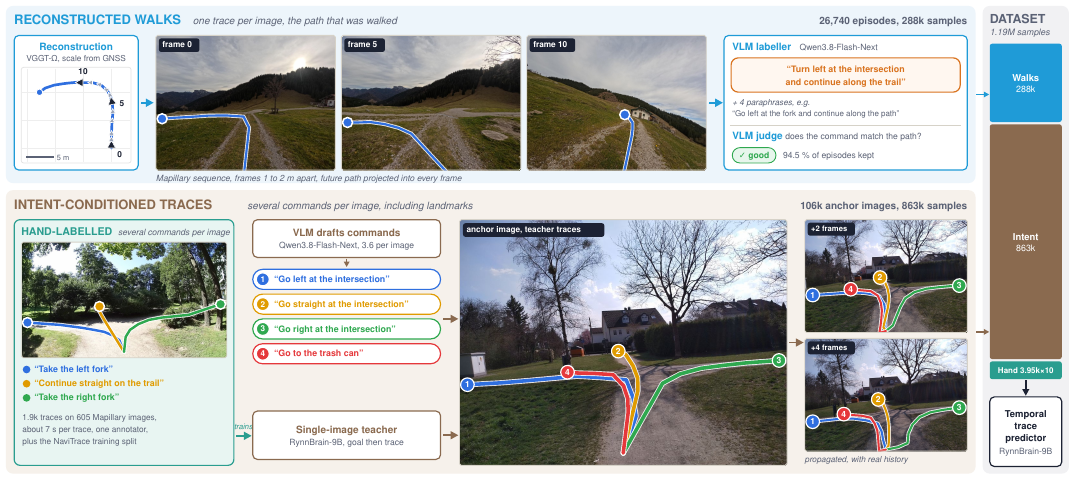}
\par\vspace{-7pt}%
\caption{Training data of the trace predictor. Top, reconstructed walks. A Mapillary sequence is reconstructed with
VGGT-$\Omega$ and scaled with GNSS, the future path is projected into every frame, a VLM labeller writes the atomic
command with four paraphrases, and a VLM judge keeps only consistent episodes. Every image carries a single trace, the
path that was walked. Bottom, intent-conditioned traces. Hand-labelled traces train a single-image teacher, a VLM drafts
several commands for an anchor image, the teacher draws one trace per command, and the reconstruction propagates the
traces to subsequent frames, which provides the real history. Right, composition of the 1.19M training samples.}
\label{fig:dataengine}
\end{figure*}

Training requires samples $(\vect{x}, c, \boldsymbol{\tau})$ whose trace follows the command. Projecting the future path
of walking sequences into earlier frames yields traces at
scale~\cite{liu2025citywalker,hirose2024lelan,mei2026urbannav,seneviratne2026vega}, but the intent of the walker is
unknown, many crowd-sourced sequences have no clear destination, and every image comes with a single trace, namely the
path that was walked. A model trained only on such data can associate an image with a path while ignoring the command,
and it never sees the branch that was not taken or a path towards an object. We therefore combine reconstructed walks,
which provide real temporal context, with intent-conditioned traces, which pair the same image with several commands,
for example ``Take the left fork'' and ``Take the right fork'' (\cref{fig:dataengine}). Both are built from
crowd-sourced street-level sequences~\cite{mapillary} (51,907 sequences from 182 cities), and the test sequences are
excluded (\cref{app:data}).

\textbf{Reconstructed walks.} We select episodes at junctions, turns and straight segments, reconstruct each sequence
with VGGT-$\Omega$~\cite{wang2026vggtomega}, scale it with the GNSS arc length and project the ground below the future
camera positions into every frame. The trace is cut where a building occludes it, where it leaves the image and where
the walker turns back. Since Mapillary images are not taken at uniform intervals, consecutive frames are 1 to 2\,m
apart, which also varies the spacing of the history. A VLM writes the atomic command that the drawn path depicts
together with four paraphrases, and a second VLM pass keeps only the episodes whose command and path agree. This yields
288k samples from 26,740 episodes.

\textbf{Intent-conditioned traces.} We first train a single-image teacher $f^{\mathrm{T}}$ with the two-stage
formulation~\eqref{eq:twostage} on the training split of NaviTrace~\cite{navitrace} and on 1.9k traces that we drew by
hand for several commands on each of 605 Mapillary images, at about 7\,s of labelling effort per trace. We then mine
anchor images at junctions, crossings and trails, and a VLM drafts 3.4 commands per kept image on average, one for each
visible branch or crossing direction and up to two towards unique landmarks. The teacher predicts a trace for every
command, and the reconstruction propagates it into the subsequent frames of the sequence, which provides the history
that the teacher does not see. This yields 863k samples from 106k images. Together with the hand-labelled traces, the
dataset contains 1.19M samples. Adding the intent-conditioned samples improves the open-loop accuracy monotonically
(\cref{sec:q4}).

\subsection{Execution with an RL Navigation Policy}
\label{sec:execution}

A new trace becomes available only every 1.5 to 3\,s, and it describes the scene at the time its image was captured. In
the meantime, the robot moves on and pedestrians or vehicles may enter its path. The trace predictor also reasons about
the route rather than about the clearance of the robot, so a trace can pass close to a pole or through a parked bicycle.
The trace therefore cannot be executed open loop. It requires a low-latency controller that knows the embodiment, tracks
the trace and avoids obstacles on its own.

For this, we lift the trace to 3D with the depth image, anchor it in the odometry frame and keep only its prefix on
observed walkable ground (\cref{app:implementation}). A carrot $\vect{q}_t$ moves along the lifted path 4.5\,m ahead of
the robot and serves as the goal of an RL navigation policy~\cite{yang2025sru}. The policy is trained in simulation to
reach point goals from depth images while avoiding obstacles, and it runs at 5\,Hz on the latest depth image and the
velocity of the robot, so that the robot keeps following the previous path while a new trace is computed. A new trace is
requested on a new command, after 3\,m of travel, after 40\,\% of the trace has been executed or after 5\,s. Under the
final command, the robot stops once the goal is within 1.5\,m.

\section{Experimental Setup}
\label{sec:setup}

We evaluate the full system in closed loop in a simulation benchmark, where all methods run on the same missions and
GNSS errors, and on a real robot, and the trace predictor in open loop against reference traces.

\subsection{Closed-Loop Simulation Benchmark}
\label{sec:sim}

\begin{figure}[t]
\centering
\includegraphics[width=\columnwidth]{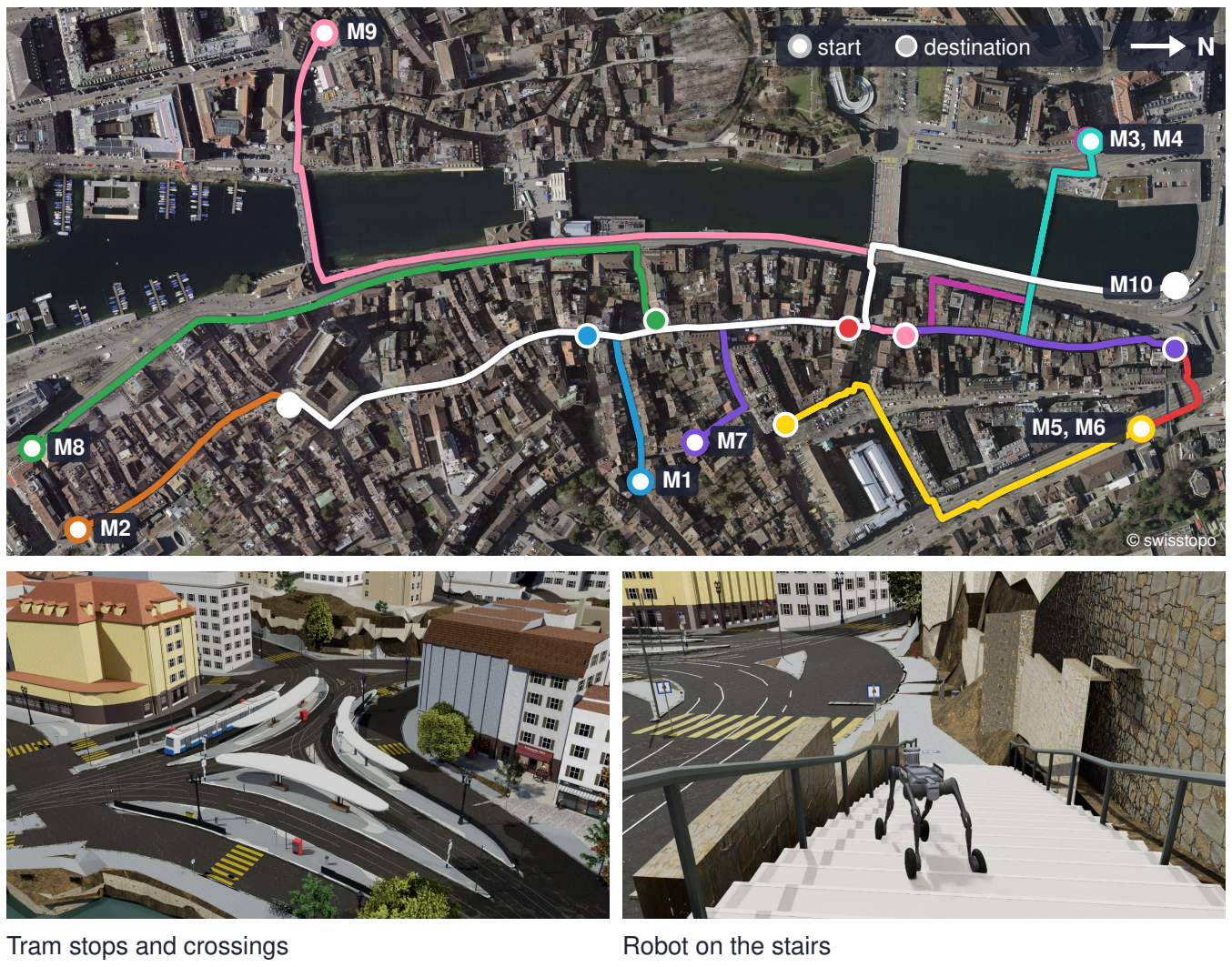}
\par\vspace{-7pt}%
\caption{Simulation benchmark. Top, the ten long-range missions in the simulated \cityname old town. Bottom, renders of the
simulated tram stops and crossings at a tram station and of the robot on a staircase.}
\label{fig:missions}
\end{figure}

We reconstruct the \cityname old town in Isaac Sim~5.0 from \anon{public government geodata}{Swiss federal geodata}, OpenStreetMap and street-level imagery
(\cref{fig:missions}). The static scene covers 0.91\,km$^2$ with 123 zebra crossings and 880 shops and restaurants, so
that missions can end at named destinations. The robot is a simulated Unitree B2W wheeled-legged robot. The benchmark
consists of ten missions between 152 and 864\,m, 4.68\,km in total, with 14 junction turns, 2 forks, 14 road crossings
and one staircase (\cref{app:missions}). GNSS errors are emulated by a slowly drifting bias with a radial RMS of 2, 5 or
7\,m, and each method runs three seeds per mission, giving 30 runs per condition.

\textbf{Metrics.} A run succeeds when the robot stops within 5\,m of the destination and crosses every road at the
zebra crossing, that is, at least half of its path over the road at each crossing of the route lies on the zebra. We
additionally report the crosswalk violation, the share of crossings taken outside the zebra crossing, and the road
violation, the share of the driven path on which the robot is on the road
instead of the sidewalk.
\label{sec:metrics}

\subsection{Closed-Loop Baselines}
\label{sec:baselines}

We compare against several state-of-the-art methods with publicly available models. The point-goal navigation methods
CityWalker~\cite{liu2025citywalker}, VAMOS~\cite{castro2025vamos} and OmniVLA~\cite{hirose2025omnivla} follow the route
as GNSS waypoints with their released control loops. To compare path prediction independently of the controller, we also
pair CityWalker and VAMOS with our lifting and RL policy. The language-conditioned methods
TIC-VLA~\cite{huang2026ticvla}, Foresight~\cite{zhang2026foresight} and VENTURA~\cite{zhang2025ventura} completed no
mission with our orchestrator VLM, so they receive their commands from an oracle orchestrator that issues each command
at a predefined position along the true route, tuned for each method. To separate the errors of the orchestrator from
those of the rest of the hierarchy, we also run \method with such an oracle (\cref{app:missions}).

\subsection{Open-Loop Trace Prediction}
\label{sec:openloop}

Each sample consists of an atomic command, the current image with its history and walked-path map, and a reference
trace. We use two test sets of Mapillary images from sequences excluded from training (\cref{app:results}). Human-Test
contains 953 hand-drawn traces on 332 images with several commands per image, including landmarks. Walks-Test contains
31,012 samples, all frames of 2,869 held-out episodes of reconstructed walks, labelled by the same automatic pipeline as
the training data.

\textbf{Metrics.} We report the NaviTrace (NT) score~\cite{navitrace}, a metric for image-space traces that is aligned
with human preferences. It combines the dynamic time warping distance to the reference trace, the final displacement
error and a semantic penalty for surfaces that should be avoided. The score is normalised so that a perfect trace scores
100 and a straight vertical trace through the image centre scores 0 on average. We further report the average
displacement error (ADE) in percent of the image size and the side accuracy, the share of turns whose predicted endpoint
lies on the correct side of the image. VAMOS and CityWalker cannot follow language commands, so they receive the goal
predicted by our model, which isolates the shape of their paths from the choice of the goal. Some methods plan only a
few metres ahead, Foresight for example about 5\,m, whereas the references extend to where the path leaves the view.
Scoring such a short plan against the full reference penalises it for the part it never claimed to cover. We therefore
report every metric against the full reference (Full) and against the reference cut at the horizon of Foresight, where
every prediction is cut after the same distance along the reference (Hor.).

\subsection{Real-World Setup}
\label{sec:realsetup}

We deploy \method on a Unitree B2W with a Stereolabs ZED~X stereo camera, lidar-inertial odometry~\cite{chen2023dlio}
and a GNSS receiver. Holistic Fusion~\cite{nubert2025holisticfusion} fuses the lidar-inertial odometry with the GNSS
measurements for state estimation. The trace predictor runs onboard on an NVIDIA Jetson AGX Thor, the orchestrator VLM
on a remote server, and the RL navigation policy~\cite{yang2025sru} runs on an NVIDIA Jetson AGX Orin.

\subsection{Implementation Details}
\label{sec:impl}

The orchestrator is Qwen3.8-Flash-Next~\cite{qiu2026qwen38next} with FP8 weights, used without fine-tuning. It runs on a
remote server with four H100 GPUs and takes about 1.3\,s per decision. The trace predictor is trained with LoRA adapters
on RynnBrain-9B for one epoch on the 1.19M samples. On Human-Test, the single-call model reaches a NaviTrace score of
83.2, on par with the two-stage model in \cref{tab:openloop}, and with FP8 weights it takes about 1.5\,s per request on the Jetson
AGX Thor. The open-loop results use the two-stage model, and all closed-loop experiments use the single-call model
(\cref{app:implementation}).

\section{Experiments}
\label{sec:experiments}

Our experiments examine whether expressing a web-map route as a sequence of atomic language commands, each executed as
an image-space trace, enables a robot to reach destinations in a city reliably, and what each level of the hierarchy
contributes.
We structure them around five questions.
\begin{enumerate}
  \item[\textbf{Q1}] Does the hierarchy follow web-map directions over long routes, and how do existing navigation models perform on the same task? (\cref{sec:q1})
  \item[\textbf{Q2}] How robust is it to GNSS errors, and where do its failures come from? (\cref{sec:q2})
  \item[\textbf{Q3}] Does it transfer to the real world? (\cref{sec:q3})
  \item[\textbf{Q4}] What does each component contribute? (\cref{sec:q4})
  \item[\textbf{Q5}] Which new capabilities does a language interface between map and motion enable? (\cref{sec:q5})
\end{enumerate}

\subsection{Q1: Long-Range Navigation from Web-Map Directions}
\label{sec:q1}

\begin{table}[t]
\centering
\caption{Long-range navigation with perfect GNSS. Success rate (SR) over the ten simulated missions with three seeds
each.}
\label{tab:main_sr}
\footnotesize
\setlength{\tabcolsep}{3pt}
\begin{tabular}{l l l c}
\toprule
Method & Guidance & Low level & SR [\%]\,\up \\
\midrule
CityWalker~\cite{liu2025citywalker} & GNSS waypoints & released policy & 20.0 \\
VAMOS~\cite{castro2025vamos} & GNSS waypoints & released policy & 13.3 \\
OmniVLA~\cite{hirose2025omnivla} & GNSS waypoints & released policy & 0.0 \\
\midrule
TIC-VLA~\cite{huang2026ticvla} & tuned oracle & released policy & 6.7 \\
Foresight~\cite{zhang2026foresight} & tuned oracle & path $\rightarrow$ RL policy & 3.3 \\
VENTURA~\cite{zhang2025ventura} & tuned oracle & path $\rightarrow$ RL policy & 6.7 \\
\midrule
\textbf{\method} & orchestrator VLM & trace $\rightarrow$ RL policy & \textbf{83.3} \\
\bottomrule
\end{tabular}
\end{table}

We first evaluate whether the full system follows web-map directions over complete missions. Each method runs the ten
missions of the simulation benchmark (\cref{sec:sim}), three times each with perfect GNSS. \method completes 83.3\,\% of
the missions, whereas no existing method exceeds 20\,\% (\cref{tab:main_sr}). The point-goal methods follow GNSS
waypoints along the route with their released control loops (\cref{app:missions}). They fail mostly by stalling at
obstacles along the sidewalk or by falling, most often at a river embankment that is the most frequent failure location
across all baselines (\cref{app:results}). OmniVLA, which receives the active waypoint as its goal pose, does not
complete any mission and collides with facades, advertising boards and street furniture.

The language-conditioned methods require a language instruction for every route segment. Driven by our orchestrator VLM,
they did not complete any mission, so we provide them with the strongest available instruction source instead. An oracle
orchestrator issues each route command from the true pose without latency, and its command timing and wording are tuned
separately for each method (\cref{sec:setup}). Even under these conditions, the language-conditioned methods rarely
succeed. All of their successes occur on the shortest mission (152\,m), and on the remaining nine missions they leave
the route within the first 10 to 130\,m. TIC-VLA over-rotates on turn commands and drifts off straight sidewalks, while
Foresight and VENTURA turn into walls at junctions or follow side paths. These results indicate that executing
individual atomic instructions reliably is the limiting factor for these models, which we examine further in
\cref{sec:q4}.

\subsection{Q2: Robustness to GNSS Errors and Failure Attribution}
\label{sec:q2}

\begin{figure*}[t]
\centering
\includegraphics[width=\textwidth]{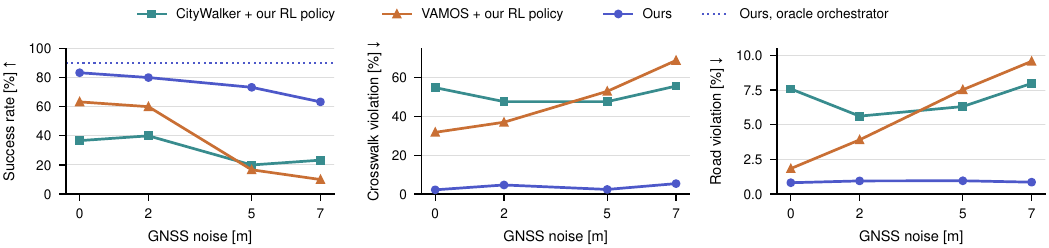}
\par\vspace{-7pt}%
\caption{Long-range navigation under GNSS noise. Success rate, crosswalk violation and road violation of \method and of
the point-goal baselines with our RL navigation policy. The dotted line in the success-rate panel marks \method with the oracle orchestrator at perfect GNSS.}
\label{fig:gps_noise}
\end{figure*}

CityWalker predicts waypoints learned from city walking videos, and VAMOS re-ranks candidate paths with an
embodiment-aware affordance model. On our robot, however, several of their runs fail at the low level, by stalling at
obstacles or by falling (\cref{sec:q1}). Paths predicted from images do not reliably keep the clearance that the robot
needs, which also holds for the traces of \method (\cref{sec:execution}). A low-latency policy that avoids obstacles is
therefore an integral part of the hierarchy of \method, and for a comparison on the same low level, we replace the
controllers of CityWalker and VAMOS with this RL navigation policy, which tracks their predicted paths through the same
lifting and carrot. With perfect GNSS and our RL policy, these variants rarely stall or fall, but they cross the road outside the
zebra crossing at 55\,\% and 32\,\% of the crossings, since a GNSS waypoint does not indicate where the crosswalk lies. They
therefore succeed in only 36.7\,\% and 63.3\,\% of the runs, compared with 83.3\,\% for \method. We then corrupt the position given to every
method with a drifting GNSS error of 2, 5 and 7\,m RMS (\cref{fig:gps_noise}). As the GNSS error grows, their success rate
drops to 23.3\,\% and 10.0\,\% at 7\,m. The additional failures are falls at kerbs and embankments onto which displaced
waypoints lead the robot, stops short of or beyond the destination, and stalls at obstacles (\cref{app:results}). In
contrast, the success rate of \method decreases from 83.3\,\% to 63.3\,\%, and its crosswalk and road violations remain
below 6\,\% and 1\,\%, respectively, at every noise level. Across all noise levels, \method leaves the crosswalk or the
sidewalk about ten times less often than the baselines. This is because \method predicts the path in the camera image
rather than from the GNSS position, so GNSS errors do not shift the robot off the sidewalk or the crosswalk. Instead,
they affect the timing of the orchestrator's decisions, which relies on the distance to the next manoeuvre, and most
failures of \method occur when a manoeuvre is commanded too early, too late or at the wrong junction, so that the robot
leaves the planned route.

To measure how much the orchestrator limits \method, we replace the orchestrator VLM with the oracle orchestrator while
keeping the trace predictor and the RL policy unchanged. The success rate rises from 83.3\,\% to 90.0\,\%.

\subsection{Q3: Real-World Deployment}
\label{sec:q3}
\begin{table*}[t]
\centering
\caption{Real-world missions of \method in \cityname. Length of the planned route, mean speed from the start of motion to the stop at
the destination, number of commands issued by the orchestrator, number of operator interventions and share of the driven
distance without intervention.}
\label{tab:realworld}
\footnotesize
\setlength{\tabcolsep}{2.9pt}
\begin{tabular}{@{}lllccccc@{}}
\toprule
Mission & Type & Characteristics & Length [m] & Speed [m/s] & Commands & Interventions & Autonomy [\%] \\
\midrule
Starbucks & city centre & stairs, crosswalks, tram tracks, construction sites & 426 & 1.12 & 26 & 0 & 100 \\
Snipes & city centre & crowded streets, construction site, crosswalk & 248 & 1.01 & 13 & 0 & 100 \\
Dieci & city centre & footbridge, tram crossing, sidewalks, crowded streets & 312 & 1.11 & 19 & 0 & 100 \\
Mendokoro & rural trail, campus & road, field paths, forest trails, plaza, ramp & 1368 & 1.19 & 21 & 2 & 97.2 \\
\bottomrule
\end{tabular}
\end{table*}

\begin{figure*}[t]
\centering
\includegraphics[width=\textwidth]{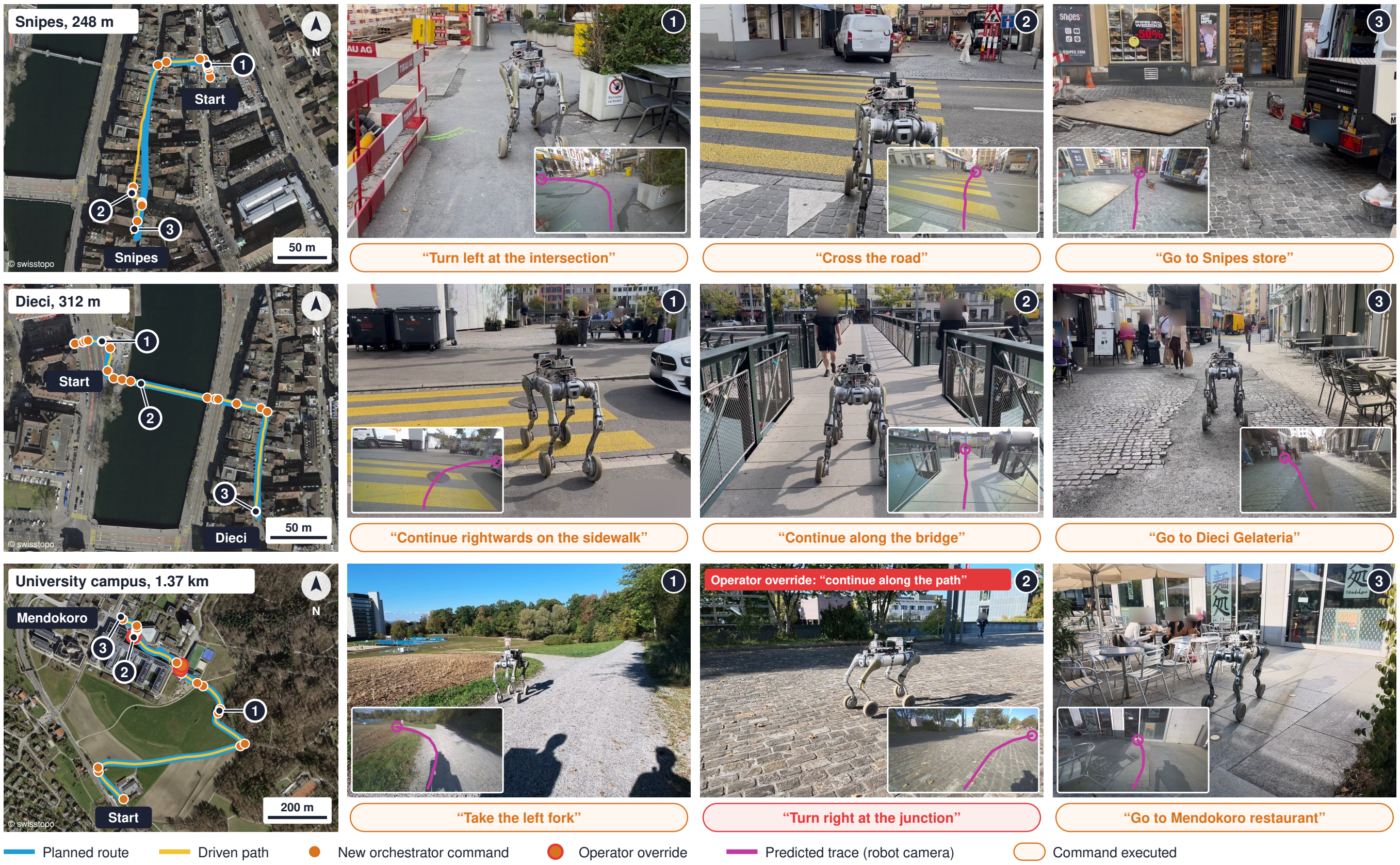}
\par\vspace{-7pt}%
\caption{Three further real-world missions in \cityname. Left, the planned route, the driven path, new orchestrator
commands and operator overrides. Right, three moments per mission with the robot camera and the predicted trace (inset)
and the executed command.\imagery}
\label{fig:realworld}
\end{figure*}

We deploy \method on a Unitree B2W wheeled-legged robot in \cityname on four missions between 248\,m and 1.37\,km
(\cref{tab:realworld}, \cref{fig:teaser,fig:realworld}). The routes come from the web map, and the same models, prompts
and parameters as in simulation are used, apart from the RL policy, which is retrained for the real robot.

The Starbucks mission leads from a university building to the nearest Starbucks in the city centre over 426\,m
(\cref{fig:teaser}). The robot crosses two streets with tram tracks, passes a roadside construction, descends two
sets of stairs and crosses the wide roads on the zebra crossing before it stops next to the A-board of the Starbucks. At one tram crossing, the operator briefly paused the robot to
check for an approaching tram, since \method cannot yet react to trams (\cref{sec:discussion}). We do not count this
pause as an intervention, because the commands and the path were not changed.
The Snipes and Dieci missions lead through the old town, where the robot shares crowded pedestrian streets with
pedestrians and restaurant tables, passes construction sites and, in the Dieci mission, crosses a footbridge and a
crossing over tram tracks (\cref{fig:realworld}). In such places, the walkable space is often only a few metres wide,
for example the passage between the construction barriers or the footbridge, so a waypoint displaced by a few metres of
GNSS error would already lie in the construction site or off the bridge. \method instead draws the path in the camera
image along the visible passage. In the Snipes mission, the position estimate drifts by up to 12.7\,m from the route
while the robot approaches a crosswalk, yet the robot crosses on the zebra, because the crossing is executed using the
context from the robot camera images (\cref{fig:realworld}, top, moment~2).

The Mendokoro mission covers 1.37\,km on a road, field paths, forest trails and a university campus. On this mission,
the operator intervened twice by replacing the current command with an alternative language command, without
teleoperating the robot, after turns at the campus entrance and on an open plaza (\cref{fig:realworld}, moment~2). In
both cases, the orchestrator did not switch back to a straight command after the turn had been taken, and it kept or
reissued a turn command instead. On an open plaza, the camera image does not show the intersection that the turn refers
to, so the trace predictor continues to draw a turning path, and the robot would turn around if the orchestrator did not
issue the straight command in time. The operator therefore sent ``continue straight along the path'', after which the
orchestrator resumed. This failure shows a limitation of commands that refer to features that are no longer visible,
which we discuss in \cref{sec:discussion}.

\subsection{Q4: Component Analysis}
\label{sec:q4}

\begin{table*}[t]
\centering
\caption{Open-loop trace prediction on the two test sets (\cref{sec:openloop}), scored against the full reference (Full)
and the reference cut at Foresight's horizon (Hor.). Best per column in bold. $^\dagger$Given our predicted goal.}
\label{tab:openloop}
\footnotesize
\setlength{\tabcolsep}{4.5pt}
\begin{tabular}{l *{12}{c}}
\toprule
& \multicolumn{6}{c}{Human-Test (hand-labelled, 953 samples)} &
  \multicolumn{6}{c}{Walks-Test (auto-labelled, 31,012 samples)} \\
\cmidrule(lr){2-7}\cmidrule(lr){8-13}
& \multicolumn{2}{c}{NT\,\up} & \multicolumn{2}{c}{ADE\,\down} & \multicolumn{2}{c}{Side\,\up}
& \multicolumn{2}{c}{NT\,\up} & \multicolumn{2}{c}{ADE\,\down} & \multicolumn{2}{c}{Side\,\up} \\
\cmidrule(lr){2-3}\cmidrule(lr){4-5}\cmidrule(lr){6-7}\cmidrule(lr){8-9}\cmidrule(lr){10-11}\cmidrule(lr){12-13}
Method & Full & Hor. & Full & Hor. & Full & Hor. & Full & Hor. & Full & Hor. & Full & Hor. \\
\midrule
\tabinput tables/exp1_openloop.tex
\bottomrule
\end{tabular}
\end{table*}

\begin{figure*}[t]
\centering
\includegraphics[width=\textwidth]{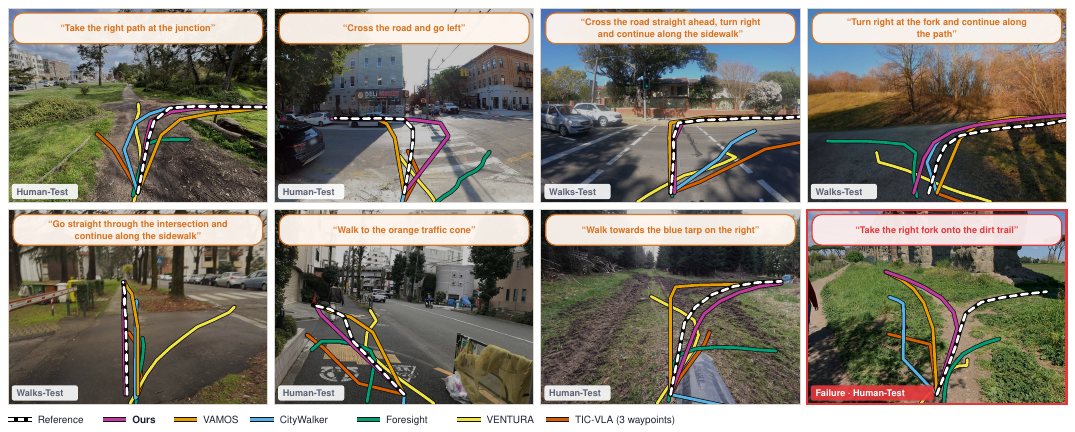}
\par\vspace{-7pt}%
\caption{Open-loop trace prediction for single atomic commands covering turns, forks, crossings, junctions and Objects.}
\label{fig:openloop}
\end{figure*}

\textbf{Trace prediction.} We first evaluate the trace predictor in open loop on the two test sets of
\cref{sec:openloop} (\cref{tab:openloop}). The trace predictor of \method reaches a NaviTrace score of 83.2 on the
hand-labelled set, compared with at most 57.0 for the baselines, and it chooses the correct side of a turn in 97.9\,\%
of the cases. \Cref{fig:openloop} shows qualitative examples, including one of the failures. The vision-language navigation models
TIC-VLA, Foresight and VENTURA choose the wrong side in a quarter to a third of the turns on the same set. This confirms
that these models are brittle already at the level of a single atomic instruction, which explains their failures over
long routes in \cref{sec:q1}.

\textbf{Temporal history.} We compare the temporal trace predictor with a model that observes only the current frame on
50 closed-loop episodes in the simulated old town, each driven by a single command with three seeds. With
history, the robot succeeds in 94.7\,\% of the runs, compared with 74.7\,\% without. The gap arises mainly from the
turns: without history, the robot keeps executing a turn after it is complete (\cref{app:hist}). The history thus lets
the trace predictor recognise the progress of the requested action, which the current frame alone cannot provide.
\begin{figure}[t]
\centering
\includegraphics{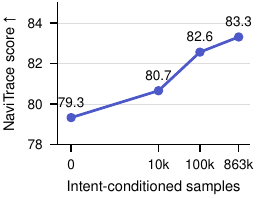}
\par\vspace{-7pt}%
\caption{NaviTrace score on Human-Test against the number of intent-conditioned samples added (logarithmic axis).}
\label{fig:datascale}
\end{figure}

\textbf{Intent-conditioned data.} Adding the intent-conditioned samples (\cref{sec:data}) improves trace prediction
monotonically, from a NaviTrace score of 79.3 with the hand-labelled samples and reconstructed walks alone to 83.3 with
863k additional samples (\cref{fig:datascale}). Reconstructed walks pair each image with the single path that was
walked, whereas the intent-conditioned samples pair the same image with several commands, including the branches that
were not taken and paths to landmarks. We attribute the improvement to this diversity of commands per scene, which
encourages the model to follow the command rather than the most likely path.

\begin{table}[t]
\centering
\caption{Orchestrator decision benchmark on 957 replayed decisions. Critical-error rate overall and with 0 and 5\,m GNSS
error, median time per call and API cost per 1,000 calls.}
\label{tab:vlm}
\footnotesize
\setlength{\tabcolsep}{3pt}
\begin{tabular}{l ccc cc}
\toprule
& \multicolumn{3}{c}{Critical [\%]\,\down} & & \\
\cmidrule(lr){2-4}
Orchestrator VLM & all & 0\,m & 5\,m & s/call & \$/1k \\
\midrule
\tabinput tables/exp4b_vlm_compact.tex
\bottomrule
\end{tabular}
\end{table}

\textbf{Orchestrator VLM.} Finally, we compare orchestrator models on single decisions replayed from the logged requests
of the long-range runs, with the same prompt, map, camera images and request text that the live system sent
(\cref{tab:vlm}). The label of each decision follows from the route and the true position of the robot. A decision is
critical when it would end the mission, for example a manoeuvre issued too early, a manoeuvre of the wrong kind or side,
a missed manoeuvre or leaving the zebra early. The self-hosted Qwen3.8-Flash-Next model used by \method makes a critical
error in 5.1\,\% of the decisions, mostly at crosswalk entries and turns. The strongest commercial models reduce this
rate to 3.2\,\%, but they take three to five times longer per call and require a commercial API. This trade-off
motivates our choice of a fast self-hosted model, and it suggests that the orchestrator will improve directly with faster
and stronger VLMs.

\subsection{Q5: Applications of the Language Interface}
\label{sec:q5}
\begin{figure*}[t]
\centering
\includegraphics[width=\textwidth]{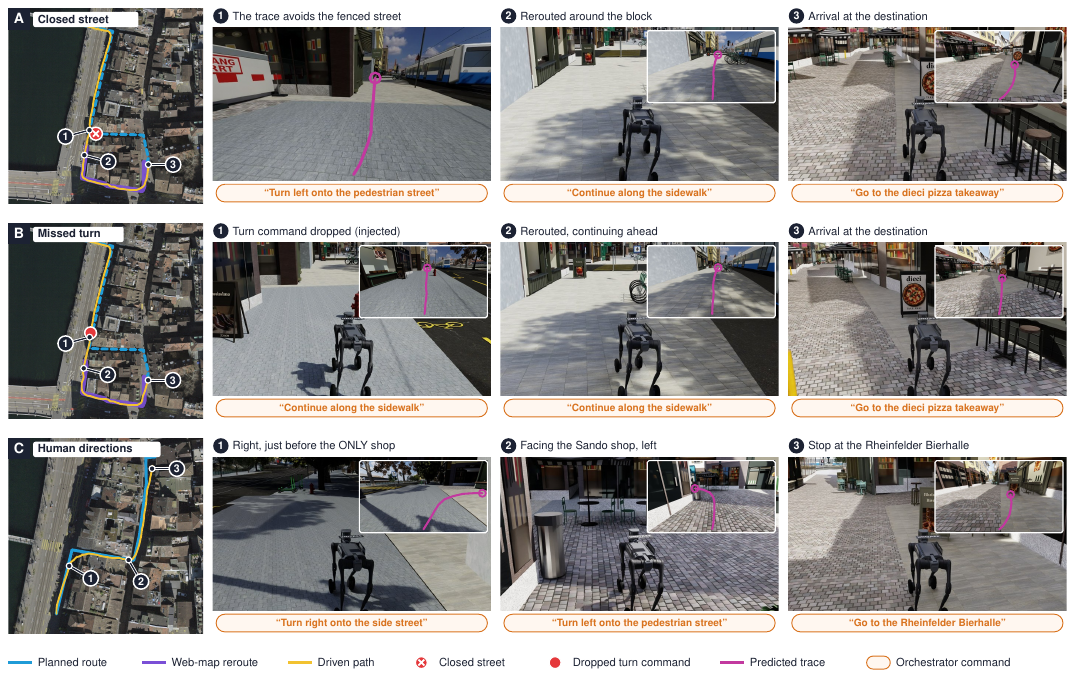}
\par\vspace{-7pt}%
\caption{Applications of the language interface in simulation: (A)~a closed street, (B)~a missed turn and (C)~directions
from a person without a map or GNSS. Numbered markers on the maps correspond to the frames, with the robot camera and
the predicted trace (inset).\imagery}
\label{fig:demos}
\end{figure*}

Beyond following a given route, the hierarchy also handles situations in which the planned route cannot be followed,
such as a closed street or a missed turn, and missions for which no route is available at all. Since the route enters the hierarchy only through language commands, these cases are handled at the level of the directions, without changes to the
trace predictor or the RL policy. We demonstrate three such cases in simulation (\cref{fig:demos}).

In the first, a construction fence closes a street on the planned route. When the orchestrator commands the turn into
this street, the trace predictor does not draw a path into the fence, and the robot continues along the sidewalk
(\cref{fig:demos}, frame~A1). Having passed the closed street, the robot requests a new route from the web map, which
returns an alternative path around the block, and it reaches the destination along this path. In the second, we mimic an
orchestrator failure by dropping a turn command, so that the robot continues straight past the turn. As in a web map
that recomputes the route after a missed turn, the robot requests a new route once it is 10\,m off the planned route,
and it reaches the destination along the new route without turning back.

Finally, a person asked for directions typically describes the way as a sequence of steps grounded in landmarks, such as
``Turn right into the narrow lane just before the black ONLY clothes shop''. We give six such directions to the
orchestrator and run a short mission of 127\,m without any map, route or GNSS, using only the camera images and
odometry. The robot successfully navigates to the entrance of the destination using the provided directions. Further
details are given in \cref{app:implementation}.

\section{Discussion and Limitations}
\label{sec:discussion}

\textbf{Orchestrator.} The off-the-shelf orchestrator accounts for several of the failures (\cref{sec:q2}). It issues
turns too early or too late, takes the wrong junction when the target street is narrow or not yet in view, enters
crosswalks too early and sometimes does not switch back to a following command after a manoeuvre (\cref{sec:q3}). In
the Starbucks mission, for example, it issued one of the crossing commands several metres before the crosswalk, so that the robot
stepped onto the road beside the crosswalk and joined it only partway across.
Moreover, we observe that it tends to weigh the text-based distance to the next manoeuvre more than the camera images.
Since this distance inherits the GNSS error, this might explain the decrease of the success rate with the GNSS error,
while the rule violations remain low. Stronger commercial models reduce critical decisions by about a third, but at the
cost of a three to five times higher latency (\cref{tab:vlm}). Fine-tuning the orchestrator VLM on decisions that are
verified by executing them with the trace predictor is a promising direction. The orchestrator also relies on a few
heuristic rules on the robot and on a network connection to a remote server.

\textbf{Dependence on GNSS and on the web map.} GNSS only identifies the current step of the route, which makes \method
robust to drifting errors of several metres, but long dropouts leave the distance to the next manoeuvre stale. The
method also inherits the limitations of the web map, such as routes that omit a crossing.

\textbf{Scaling to longer missions.} The hierarchy places no limit on the mission length, since any route decomposes
into atomic manoeuvres. With a critical-error rate of about 5\,\% per decision (\cref{tab:vlm}), however, a route of
several kilometres with tens of turns and crossings would have a high failure rate, so the reliability of the
components, in particular of the orchestrator, has to improve before such missions become practical.

\textbf{Trace predictor.} The image history covers only about 9\,m of travel, whereas crossing a wide road takes 20 to 30\,s,
so long manoeuvres depend on the orchestrator issuing entry and exit commands at the right time. Since the walked paths
in the training data pass around obstacles that the projected traces cross, the trace predictor also draws traces
through obstacles and relies on the lifting and the RL policy to avoid them. Finally, it sees only the front camera, so
in open areas without visible context, such as the plaza in the Mendokoro mission (\cref{sec:q3}), it can keep turning
after a turn is complete.

\textbf{Latency and the RL policy.} A new command takes effect in the executed path about 3\,s after its image (1.3\,s
for the orchestrator VLM and 1.5\,s for the trace predictor), about 3\,m at 1\,m/s, which can be enough to miss a
narrow opening. This is tolerable only because the RL policy runs at 5\,Hz
on the latest depth image, keeps following the previous path and avoids obstacles and pedestrians. Moreover, because of
this latency, safety-critical reactions such as stopping at a red light or for an approaching tram cannot currently be
handled. Lower latency would also allow speeds above the current maximum speed cap of 1.3\,m/s.

\textbf{Evaluation and other embodiments.} Our metrics measure compliance with pedestrian rules, such as crossing at
crosswalks, but not the interaction with traffic, and the simulation lacks traffic and pedestrians. Moreover, the real-world missions are
qualitative demonstrations without baselines. In principle, the hierarchy transfers to other platforms, such as humanoid
robots, by replacing only the RL policy, possibly together with an embodiment-aware trace predictor~\cite{navitrace},
which remains future work.

\section{Conclusion}
\label{sec:conclusion}

We present \method, a hierarchy that follows web-map directions through a city by translating them into atomic language
commands and executing each command as a trace in the camera image, which an embodiment-aware RL policy tracks.
Separating the decision of which command applies
from its execution makes the system robust to GNSS errors and keeps the robot on crosswalks and sidewalks, where
point-goal methods fail. A temporal trace predictor trained on automatically labelled street-level data executes atomic
commands far more reliably than existing vision-language navigation models. Many of the remaining failures stem from the
orchestrator, which makes improving its decisions the main direction for future work. Real-world missions of up to
1.37\,km show that the approach transfers from simulation to city streets. We will release the simulation, the
benchmark, the models, the dataset and the code to support further research on long-range urban navigation.

\ifanonymous\else
  \section*{Acknowledgment}
  The authors would like to thank Turcan Tuna for help with the state estimation and Claudio Bonaglia, Tim Tschanz and
  Cyril P\"{u}ntener for help with the real-world deployments. This work was supported by the NVIDIA Academic Grant Program through the project Hierarchical Vision-Language
  Navigation Foundation Model. It was also supported as part of the Swiss AI Initiative by a grant from the Swiss
  National Supercomputing Centre (CSCS) under project ID abs19 on Alps, by the Luxembourg National Research Fund
  (Ref.~18990533) and by the Swiss National Science Foundation (SNSF) through projects No.~200021E\_229503 and
  No.~227617.

  \section*{Author Contributions}
  MP led the project, developed the method, built the data pipeline and the simulation benchmark, carried out the
  experiments and the real-world deployments and wrote the manuscript. JF contributed to the brainstorming and the
  development of ideas throughout the project. PR trained the RL navigation policy used in the real-world deployments and
  helped with the deployments. MH and CC supervised the project. All authors reviewed and edited the manuscript.
\fi

\balance
\bibliographystyle{IEEEtran}
\bibliography{references}

\clearpage
\nobalance
\appendices
\crefalias{section}{appendix}\crefalias{subsection}{subappendix}   %

\noindent\textbf{Use of generative AI.}  A large language model assistant was\ used
to edit the text of all sections and to write the scripts that generate the figures and tables from the logged
experiments. The authors reviewed all content and take full responsibility for it.

\section{Implementation Details}
\label{app:implementation}

\subsection{Route}
The destination is resolved with the Google Places API, and the Google Routes API provides a walking route from the
current position. The route is replanned on the pedestrian network of OpenStreetMap and divided into steps at
manoeuvres, crossings, stairs and changes of the street name.

\subsection{Orchestrator Request}
Each request contains the destination, the directions within 20\,m of the current route position, the next direction
with its approximate distance, the distance of the GNSS position from the planned route, the current command, a
satellite image of 50\,m width rotated heading-up and up to five camera images. The directions are listed with one of
three states, namely ``in about $N$\,m'', ``done, $N$\,m behind'' or ``in progress, far kerb in about $N$\,m''. An
example request is given in \cref{app:prompts}. The camera images are selected by the distance travelled along the pose
history, 3, 6, 9, 12 and 15\,m before the current position. Images are sent as JPEG with at most 640 pixels on the long
side. The model answers in JSON with the reasoning first and the command second. The orchestrator is Qwen3.8-Flash-Next
with FP8 weights, served with vLLM on four H100 GPUs with tensor parallelism, temperature 0 and thinking disabled. A new
request is sent every 2\,s with at most one request in flight. In simulation, the answer is applied 1.2\,s after the
capture of its images, which matches the median latency of the server (1.21 to 1.29\,s across the campaigns).

The prompt prescribes a fixed sequence of commands at a road crossing, namely an entry command (``Go left onto the
crosswalk to the far end kerb''), the crossing itself (``Cross the road'') and an exit command that already contains the
next turn (``Continue rightwards on the sidewalk''). The final command ``Go to the \textless destination\textgreater''
is issued once the destination is visible, and no further orchestrator calls are made after it has been accepted.

\subsection{Rules on the Robot}
\Cref{tab:hostrules} lists the rules that guard the orchestrator's decisions.

\begin{table}[h]
\centering
\caption{Rules around the orchestrator in simulation and on the robot.}
\label{tab:hostrules}
\footnotesize
\setlength{\tabcolsep}{3pt}
\begin{tabular}{p{2.3cm} p{5.8cm}}
\toprule
Rule & Behaviour \\
\midrule
Crossing rejection & Rejects a crossing command while the next crossing entry is more than 10\,m ahead or on the other
side. \\
Arrival rejection &
  Rejects ``Go to the \textless destination\textgreater'' while the destination is more than 15\,m ahead. \\
Arrival lock & No further calls once the final command is accepted. \\
Turn display & A passed turn stays the next direction until the heading has turned by up to 45$^\circ$ or the robot is
15\,m past it. \\
Command expiry &
  A command expires 12\,s after its image capture, and the robot holds after 20\,s without a valid command. \\
Off-route & More than 20\,m from the route ends the run in simulation and pauses the robot. \\
\bottomrule
\end{tabular}
\end{table}

\subsection{Trace Predictor}
Both stages of the two-stage model are LoRA adapters of rank 64, trained with a learning rate of $4\cdot10^{-5}$ and
effective batch sizes of 256 and 192, which took about 700 GPU hours on 16 GH200 GPUs. The single-call model is trained
on 250k samples drawn from the dataset in proportion to its sources, with $\lambda_g = 2$ and $\lambda_{\mathrm{CE}} =
0.5$ in \eqref{eq:head}. In 30\,\% of these samples, the hidden state is computed with the model's own predicted goal
instead of the reference goal, so that the head learns to complete a slightly wrong goal into a consistent trace. FP8
quantisation changes the NaviTrace score on Human-Test by 0.02. Both models are trained by supervised fine-tuning only
(\cref{app:teacher}). The trace predictor receives five past images with 448 pixels on the long side, the current image
with 640 pixels on the long side and a $448\times448$ top-down map of the walked path within $\pm16$\,m. The images
precede the command in the prompt. The MLP head consists of a LayerNorm and two linear layers
($4096\rightarrow1024\rightarrow20$) with a GELU activation and reads the hidden state after the end of the goal. In the
30\,\% of the samples that use the model's own goal, the cross-entropy on the goal tokens is masked. The head model is
trained with learning rates of $3\cdot10^{-3}$ for the head and $2\cdot10^{-4}$ for the adapter, which took about 53 GPU
hours. Quantisation uses dynamic FP8 with per-channel weights and per-token activations. The vision encoder and the
output layer stay in bf16. On the Jetson AGX Thor, the median request takes 1.48\,s, and the field logs show medians of
1.21 to 1.44\,s.

\subsection{Execution}
Depth is taken from the simulator or from the stereo camera of the robot. Each trace point is lifted with the median of
the valid depth (0.3 to 25\,m) in a $5\times5$ window, or by intersecting its ray with the ground plane when no depth is
available. The ground filter fits a plane to the depth points in a $15\times15$ pixel window around each trace point and
rejects it when the residual exceeds 0.12\,m, the normal deviates by more than 40$^\circ$ from the vertical, or the
height differs from the running ground plane by more than 0.4\,m plus 0.03\,m per metre of distance. Up to two
consecutive rejected points, for example at a pole or a kerb, are bridged if the trace continues on the ground behind
them. If the trace is cut at an obstacle, its last 0.5\,m is removed so that the robot stops short of the obstacle. The
carrot is placed 4.5\,m ahead along the path and published at 5\,Hz. A new trace is requested on a new command, after
3\,m of travel, after 40\,\% of the trace has been executed or after 5\,s. In simulation, the RL policy is the
point-goal RL navigation policy of~\cite{yang2025sru} with a velocity scale of 0.7, which yields a mean speed of about
0.95\,m/s. On the robot, a retrained version of this policy is used.

\subsection{Directions from a Person}
In the mission with directions from a person (\cref{sec:q5}), the orchestrator receives the six directions, the current
command, the camera images of the last 15\,m and a line from odometry that reports the turns completed so far and the
distance walked since the last turn, but no map, route, GNSS position or distances. A few rules on the robot use
odometry only. A turn command stays active until the heading has changed by at least 50$^\circ$ in that direction or the
robot has walked 18\,m, no new turn is accepted within 10\,m of the last completed one, and the speed is limited to
0.6\,m/s. The prompt asks the orchestrator to issue a turn when its landmark is 5 to 8\,m ahead.

\section{Training Data and Teacher}
\label{app:data}

\textbf{Source pool and selection.} The Mapillary pool contains 51,907 sequences with 15.7M images from 182 cities and
regions, of which 39,931 sequences are recorded at walking pace. Episodes span eleven frames and are selected around
junction turns verified against OpenStreetMap, geometric turns away from junctions, passages straight through junctions,
which serve as negatives for turning, and straight segments. We keep perspective cameras with a horizontal field of view
of at most 80$^\circ$ and do not undistort the images. Junction events require an OpenStreetMap node with at least three
walkable ways within 10\,m. Geometric turns require a heading change of 30 to 140$^\circ$ over 5\,m chords that agrees
with the camera orientation of the reconstruction. Sequences in which the walker steps onto a major road, pauses or
jumps are removed, and every city contributes at most 8\,\% of the events.

\textbf{Reconstruction and projection.} Each sequence is thinned to frames at least 1\,m apart along the GNSS track
(median 1.4\,m) and reconstructed with VGGT-$\Omega$ in windows of 30 frames with a stride of 10. This yields the camera
centre $\vect{C}_j$, pose $T_j$ and intrinsics $K_j$ of every frame $j$, and the metric scale of each window is the
ratio of the GNSS arc length to the reconstructed arc length. The ground point below each camera is $\vect{X}_j =
\vect{C}_j - h\,\vect{n}$, with $h = 1.5$\,m and the mean up direction $\vect{n}$ of the window, and the trace of frame
$t$ consists of the projections $\Pi(K_t, T_t, \vect{X}_j)$ of the future ground points $j > t$. The trace is cut at the
first occlusion, where it leaves the field of view, in which case it follows the image border in the direction of the
turn, where the walker turns back or at the end of the window, and it is resampled to ten points. The history of a
sample consists of the five preceding frames of the episode. Overlapping windows are aligned by their camera centres,
and episodes whose windows disagree in turning direction or by more than 3\,m are removed, as are episodes with GNSS
freezes, out-and-back paths or a first trace point more than 10\,m ahead. Only buildings and walls cut the trace. Poles,
vehicles and pedestrians do not, since the trace predictor is expected to draw the path through them and the lifting
removes the part behind an obstacle (\cref{sec:execution}). The traces reach a median of 28.8\,m ahead of the camera.

\textbf{Command labelling and judging.} The labeller receives the eleven frames with the future path, a top-down view of
the walked path over OpenStreetMap, the lateral displacement of the path from the reconstruction and the GNSS manoeuvres
as weak evidence, with the instruction that the drawn path takes precedence. Commands have 3 to 12 words, and words for
curvature such as ``bend'' or ``veer'' are not allowed, since a turn requires a visible choice. During training, each
sample uses the command or one of its paraphrases, and in 15\,\% of the samples the landmark variant if one exists. The
judge sees the same frames and classifies each episode as good, wrong command, ambiguous, erratic path or bad image.
Only good episodes are kept, which retains 94.5\,\% of them. The result is 26,740 episodes from 14,609 sequences in 161
regions, and every frame of an episode with a valid trace becomes a sample, 288k in total.

\textbf{Hand labelling.} One annotator drew all hand-labelled traces with a web tool that shows the image, the command
and, in later rounds, a proposed trace that can be accepted, edited or redrawn. The tool records the time from showing a
command to saving the trace. Counting traces that took more than 15\,s as breaks and excluding them, the 3,311 traces of
all rounds took 5.9\,h, with a median of about 7\,s per trace. The training set of the teacher contains 1,917 traces on
605 images. The test set Human-Test was labelled in the same way and is excluded from all training data.

\textbf{Anchor images.} The anchor images face a decision point on
OpenStreetMap, namely a junction, a crossing or a trail. All anchors at trail forks, junctions between streets and
trails, trail segments and crossings are kept, while street junctions are capped at 90k and plain street segments at
15k. Each sequence contributes at most five anchors and each city at most 8\,\%. Sequences of the test sets, the
hand-labelled sequences and duplicate routes are excluded.

\textbf{Intent-conditioned commands.} The commands are written by a VLM (\cref{app:cmdgen}), which receives the anchor
image and a hint about the type of place from OpenStreetMap (junction, crossing, trail or street). It rejects images in
which no command can be followed and writes one command for each visible branch from a fixed set of phrasings, such as
``Take the left fork'' or ``Cross the road and go right'', and up to two commands towards landmarks that pass a
uniqueness test. It keeps about 80\,\% of the images and writes three to five commands per kept
image, 366k in total for 106k kept images, of which about half refer to a landmark. When a teacher trace is propagated, a subsequent frame is kept
while at least three trace points that lie more than 2.5\,m ahead remain in view. The 863k samples consist of 366k
anchor samples and 497k samples in subsequent frames. The hand-labelled traces are propagated in the same way.

\textbf{Dataset.} The final dataset contains 1,190,594 samples, namely 3,950 hand-labelled and NaviTrace samples, which
are repeated ten times, 288k samples from reconstructed walks and 863k intent-conditioned samples. The test split of the
reconstructed walks consists of 10\,\% of the sequences, stratified by manoeuvre and region, and it is excluded from all
sources together with the sequences of Human-Test.

\textbf{Single-image teacher.} The teacher uses the two-stage formulation with the current image only, at 1280 pixels on
the long side. It is RynnBrain-9B trained on the training split of NaviTrace and the hand-labelled traces with
supervised fine-tuning followed by GRPO.
\label{app:teacher}

\textbf{Reinforcement learning.} We experimented with GRPO, using the NaviTrace score as reward, at several stages of
development. It improved the supervised model when the training data was small, but brought no improvement with the
large datasets, so the final trace predictor is trained with supervised fine-tuning only.

\section{Simulation Environment and Missions}
\label{app:missions}

\begin{table*}[t]
\centering
\caption{The ten long-range missions. Length of the route to follow, manoeuvres, zebra crossings on the route and time
limit in simulated seconds.}
\label{tab:missions}
\footnotesize
\setlength{\tabcolsep}{5pt}
\begin{tabular}{l l l r l c r}
\toprule
& Start & Destination & [m] & Manoeuvres & Zebras & Limit \\
\midrule
\tabinput tables/supp_missions.tex
\bottomrule
\end{tabular}
\end{table*}

The scene covers 0.91\,km$^2$ around \anon{a river}{the river Limmat}. Terrain and buildings come from \anon{the national terrain and building models}{swissALTI3D and swissBUILDINGS3D},
streets, sidewalks, crossings, shops and street furniture from OpenStreetMap, and the appearance of streets and shop
fronts from 380 Mapillary images and street-level imagery of 181 shops. Materials, props such as bicycles, e-scooters,
trams, fountains and advertising boards, and the river and bridges were added for the missions, and crossings were
repaired where the reconstructed geometry did not match reality. Each mission runs in a sub-scene that keeps all
buildings and the terrain and disables distant props more than 120\,m from the route. The routes were planned on
OpenStreetMap and checked for traversability with a collision box of $1.0\times0.8$\,m. The time limit of each mission
is 300\,s plus 2.5\,s per metre of route, and \cref{tab:missions} lists the ten missions. The robot is driven by a
learned locomotion policy and has a front camera of $1024\times576$ pixels. The scene contains no pedestrians or
vehicles. Since the simulation cannot query the web map during a mission, the directions are generated from each route
in the style of the web map. In the real world and in the rerouting experiment (\cref{sec:q5}), the route comes from
the Google Routes API, and the directions are generated from it in the same way. A run fails when the robot leaves the route by more than 20\,m, stalls, falls or exceeds the time
limit. A crossing counts as taken on the zebra when at least half of the robot's path over the road at that crossing
lies on the zebra, and crossing the road elsewhere counts as a violation. A run that reaches the destination with a
violated crossing fails. The road violation excludes stretches where the route itself follows the road, as in alleys without a sidewalk. The
routes are pedestrian routes on OpenStreetMap that were compared with the corresponding web-map walking routes, and the
positions of the generated directions were checked against the scene. The robot camera has a horizontal field of view of
104.8$^\circ$. The episodes run on a cluster of RTX~3090 GPUs at about a quarter of real time. The orchestrator and the
trace predictor are served on remote H100 GPUs, and their answers are applied 1.2 and 1.5\,s of simulation time after
the capture of their images. The crosswalk and road violations are computed from the scene labels of the surface under
the robot, sampled every 0.5\,m along the driven path. Samples at the kerb edge, with walkable ground within 0.5\,m, are
not counted as road violations.

The GNSS error is a first-order Gauss-Markov process per axis with a time constant of 45\,s and a standard deviation of
$N/\sqrt{2}$, so that the radial RMS is $N$\,m, plus white noise of $0.2N$. It affects the route position, the distances
and the map of the orchestrator. The trace predictor, the lifting and the RL policy use odometry only. Success and the
off-route criterion are evaluated with the true pose.

\textbf{Baseline protocols.} The released CityWalker and VAMOS followed waypoints on the validated route with perfect
GNSS, and the benchmark ended the run once the robot came within 3\,m of the destination. The variants with our RL
policy and OmniVLA followed the web-map walking line with their own stop at 2\,m by their GNSS position and were scored
like \method. All methods were aborted after 30\,s without motion. The baselines were also aborted after 180\,s
without 2\,m of route progress or when they were more than 40\,m from the route for 60\,s, and \method at 20\,m from
the route.
CityWalker receives five frames 1\,s apart with their positions and the active waypoint and predicts
five waypoints at 1\,Hz. VAMOS samples 50 paths in the image towards the waypoint and re-ranks them with its affordance
model. OmniVLA receives the active waypoint as a pose goal. TIC-VLA runs its released asynchronous loop with its
measured latency and its own pure-pursuit controller at up to 0.7\,m/s. Foresight (with one critique and refinement
round) and VENTURA predict a path from the current frame (Foresight with three history frames), which is executed in a
stop-and-go manner through our lifting and RL policy at up to 0.7\,m/s. Under the orchestrator VLM, TIC-VLA, Foresight
and VENTURA completed none of ten runs each. The point-goal baselines receive waypoints every 20\,m (CityWalker) or
10\,m (VAMOS, OmniVLA). Concurrent navigation foundation
models~\cite{chu2026abotn0,zhang2026qwenrobotnav,bounhar2026robostral} have not released their weights, and
LightNav-0~\cite{wang2026lightnav0} was released very recently, so we do not compare with this concurrent work.

\textbf{Oracle orchestrator for \method.} Each command is issued when the true route position passes its trigger point,
which is placed relative to the corresponding direction at the median position where the orchestrator VLM issued it in
successful runs plus 1.2\,m for its latency, for example 9.3\,m before a junction turn and 10.5\,m before a crosswalk
entry.

\section{Additional Results}
\label{app:results}

\Cref{tab:supp_plain} adds the released CityWalker and VAMOS to the comparison with perfect GNSS, and
\cref{tab:supp_failures} lists the failure causes of all methods under each GNSS condition. \Cref{tab:vlm_full} breaks
the orchestrator decision benchmark down by decision type, and \cref{tab:datascale} lists the data-scale ablation of the
trace predictor.

\begin{table}[h]
\centering
\caption{Long-range navigation with perfect GNSS, including the released point-goal baselines (30 runs each).}
\label{tab:supp_plain}
\footnotesize
\begin{tabular}{l ccc}
\toprule
& SR\,[\up] & Crosswalk viol.\,[\down] & Road viol.\,[\down] \\
Method & [\%] & [\%] & [\%] \\
\midrule
CityWalker & 20.0 & 20.8 & 1.63 \\
VAMOS & 13.3 & 12.5 & 0.47 \\
\midrule
CityWalker + RL policy & 36.7 & 54.8 & 7.58 \\
VAMOS + RL policy & 63.3 & 31.8 & 1.84 \\
\method & 83.3 & 2.2 & 0.82 \\
\bottomrule
\end{tabular}
\end{table}

\begin{table*}[t]
\centering
\caption{Failure causes per method and GNSS condition, counted over the 30 runs of each cell. Fell in water, no ground
and fell are determined from the final height of the base above the scene ground (a standing B2W is at 0.73 to 0.79\,m).
Off zebra means that the robot reached the destination but crossed a road outside the zebra crossing.
Short means that the method stopped outside the 5\,m success radius, overshot that it came within 5\,m but did not stop,
stuck that it stalled or made no progress and lost that it was aborted off the route. Each failed run has one cause, the
first match in column order.}
\label{tab:supp_failures}
\footnotesize
\setlength{\tabcolsep}{4pt}
\begin{tabular}{l c c ccccccccc c}
\toprule
Method & GNSS [m] & Runs & Off zebra & Water & No ground & Fell & Short & Overshot & Stuck & Lost & Other & Failures \\
\midrule
\tabinput tables/supp_failures.tex
\bottomrule
\end{tabular}
\end{table*}

\begin{table*}[t]
\centering
\caption{Orchestrator decision benchmark per decision type. Critical-error rate over all 957 decisions and over the
decisions replayed from the runs with 0 and 5\,m GNSS error, critical errors per decision type (follow means that the
current command should be kept, zebra means crosswalk), median time per call and API cost per 1,000 calls.}
\label{tab:vlm_full}
\footnotesize
\setlength{\tabcolsep}{4pt}
\begin{tabular}{l ccc cccccc cc}
\toprule
& \multicolumn{3}{c}{Critical [\%]\,\down} & \multicolumn{6}{c}{Critical per decision type} & & \\
\cmidrule(lr){2-4}\cmidrule(lr){5-10}
Model & all & 0\,m & 5\,m & follow & turn & zebra entry & on zebra & zebra exit & arrival & s/call & \$/1k \\
\midrule
\tabinput tables/exp4b_vlm.tex
\bottomrule
\end{tabular}
\end{table*}

\begin{table}[h]
\centering
\caption{Data-scale ablation of the trace predictor on Human-Test. All arms use the same configuration and one epoch.}
\label{tab:datascale}
\footnotesize
\setlength{\tabcolsep}{2.8pt}
\begin{tabular}{l r c c c}
\toprule
Training data & Samples & NT\,\up & ADE\,\down & Side [\%]\,\up \\
\midrule
Hand-labelled + reconstructed walks & 327.8k & 79.33 & 6.07 & 95.9 \\
+ 10k intent-conditioned & 337.8k & 80.66 & 5.67 & 97.0 \\
+ 100k intent-conditioned & 427.8k & 82.57 & 5.14 & 97.5 \\
+ 863k intent-conditioned & 1,190.6k & 83.32 & 4.94 & 98.1 \\
\bottomrule
\end{tabular}
\end{table}

\textbf{Open-loop protocol.} Human-Test contains 953 hand-drawn reference traces on 332 images, labelled like the
training traces of the teacher, of which 409 refer to a landmark. For 96\,\% of its samples, the five preceding frames
and the walked path are reconstructed from the sequence with VGGT-$\Omega$. Walks-Test contains all frames ($t = 0,
\ldots, 10$) of 2,869 judge-approved episodes from 1,524 held-out sequences in 116 regions, the held-out 10\,\% of the
reconstructed-walk sequences, with 17,069 turns, 7,996 passages straight through a junction and 5,947 straight segments.
The NaviTrace score combines dynamic time warping, the endpoint error and a semantic penalty from a Mapillary Vistas
segmentation for the human embodiment~\cite{navitrace}. ADE is the mean distance after resampling both traces to ten
points. The side accuracy is computed on samples whose reference ends more than 8\,\% of the image width from the image
centre. The NaviTrace score is computed in pixels of the original image, with the prediction resampled to the length of
the reference, and an unparseable output scores 0. VAMOS receives our predicted goal as a pixel goal, and CityWalker
receives it lifted to a metric goal on the reconstructed ground plane, with its four preceding frames and their
positions. Foresight receives the last four frames cropped to its input format, VENTURA the current frame and TIC-VLA
the current frame with four history frames, whose metric waypoints are projected through a flat-ground camera model. For
the horizon-matched columns, the reference is cut where Foresight's prediction ends (on average at 69 and 66\,\% of the
reference on the two sets), and every prediction is cut after the same distance along the reference.

\section{Temporal History}
\label{app:hist}

\begin{figure*}[t]
\centering
\includegraphics[width=\textwidth]{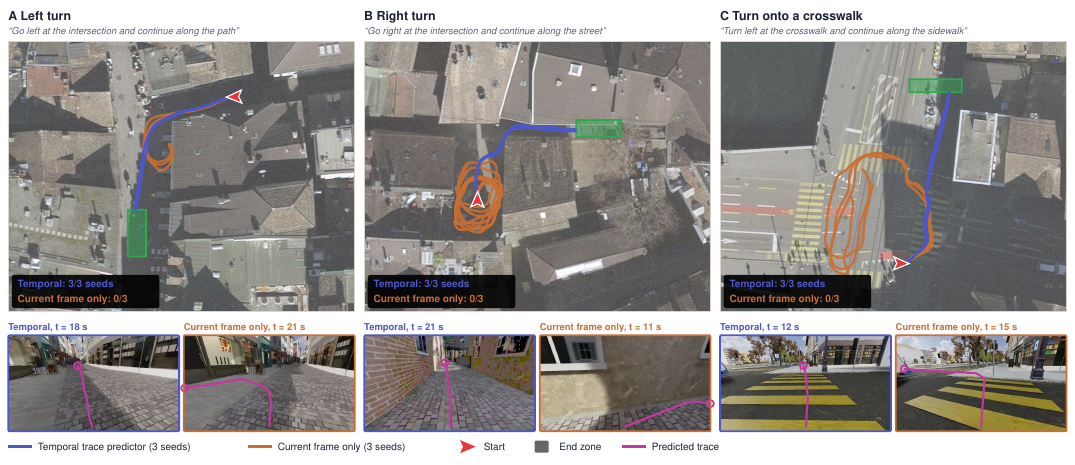}
\caption{Temporal history in closed loop. Three turn episodes in which the temporal trace predictor succeeds in all
three seeds and the model that observes only the current frame fails in all three. Top, the driven trajectories of both
models from the start (red) towards the end zone (green). Bottom, the robot camera shortly after the turn with the
predicted trace. Without history, the model keeps executing the turn after it is complete and the robot circles or turns
back, whereas the temporal model recognises that the turn has been executed and follows the new street.\imagery}
\label{fig:hist}
\end{figure*}

Each of the 50 episodes is driven by a single command, such as ``Turn left at the intersection and
continue along the street'', which the trace predictor executes for the whole episode. Half of the episodes are turns,
in which the command no longer describes the view once the manoeuvre is complete, and the other half are controls,
namely crossings, forks and straight segments. A run succeeds when the robot comes within 2\,m of an end zone on the
commanded branch. Requiring the robot to enter the end zone exactly yields 86.7 against 66.0\,\%, which preserves the
gap. \Cref{tab:hist} lists the results and \cref{fig:hist} shows typical failures of the model without history.

\begin{table}[t]
\centering
\caption{Temporal history in the trace predictor. Success rate over 50 closed-loop episodes with three seeds
each, split into turns and controls, and the share of turn runs in which the robot turned again after completing the
manoeuvre (more than 60$^\circ$ beyond the commanded turn). A run succeeds when the robot comes within 2\,m of an end
zone on the commanded branch.}
\label{tab:hist}
\footnotesize
\setlength{\tabcolsep}{3pt}
\begin{tabular}{l ccc c}
\toprule
& \multicolumn{3}{c}{SR [\%]\,\up} & Repeated turn [\%]\,\down \\
\cmidrule(lr){2-4}
Trace predictor & all & turns & controls & turns \\
\midrule
Current frame only & 74.7 & 56.0 & 93.3 & 56.0 \\
\textbf{Temporal} & \textbf{94.7} & \textbf{94.7} & \textbf{94.7} & \textbf{9.3} \\
\bottomrule
\end{tabular}
\end{table}

\onecolumn
\section{Prompts}
\label{app:prompts}

This section lists the prompts of all models, verbatim. Text in angle brackets marks images, and text in braces marks
fields that are filled in for each request.

\subsection{Orchestrator VLM}
The same prompt is used in simulation and on the robot. The model answers in JSON with a schema that enforces the fields
\texttt{reason} and \texttt{command}.

\promptfile{System prompt.}{prompts/orchestrator_system.txt} The following request is taken verbatim from a simulation
run. \promptfile{Request.}{prompts/orchestrator_user.txt}

\subsection{Trace Predictor}
The two-stage model and the single-call model use the same prompts. The single-call model receives the goal prompt and
decodes the goal, after which the MLP head reads the trace from the hidden state. The images precede the user prompt.

\promptfile{Goal stage, system prompt.}{prompts/trace_goal_system.txt} \promptfile{Goal stage, user
prompt.}{prompts/trace_goal_user.txt} \promptfile{Trace stage, system prompt.}{prompts/trace_path_system.txt}
\promptfile{Trace stage, user prompt.}{prompts/trace_path_user.txt}

\subsection{Single-Image Teacher}
The teacher uses prompts derived from the NaviTrace benchmark~\cite{navitrace} with the embodiment set to
\texttt{Human}, and it receives only the current image.

\promptfile{Goal stage, system prompt.}{prompts/teacher_goal_system.txt} \promptfile{Goal stage, user
prompt.}{prompts/teacher_goal_user.txt} \promptfile{Trace stage, system prompt.}{prompts/teacher_path_system.txt}
\promptfile{Trace stage, user prompt.}{prompts/teacher_path_user.txt}

\subsection{Command Labelling of Reconstructed Walks}
The labeller and the judge are Qwen3.8-Flash-Next, the model of the orchestrator. The frames show the future path as a red line with a yellow end dot. The
GNSS facts are listed for every turn that the GNSS track shows in the episode.

\promptfile{Labeller, system prompt.}{prompts/labeller_system.txt} \promptfile{Labeller, user
message.}{prompts/labeller_user.txt} The judge receives no GNSS facts. \promptfile{Judge, system
prompt.}{prompts/judge_system.txt} \promptfile{Judge, user message.}{prompts/judge_user.txt}

\subsection{Command Generation for Intent-Conditioned Traces}
\label{app:cmdgen}
The VLM that writes the commands is Qwen3.8-Flash-Next at temperature 0.3. It receives the anchor image and one hint that
depends on the type of place according to OpenStreetMap.

\promptfile{System prompt.}{prompts/draft_system.txt} \promptfile{User message.}{prompts/draft_user.txt}

\ifanonymous\else\fi

\end{document}